# Learning Mixtures of DAG Models

Bo Thiesson, Christopher Meek, David Maxwell Chickering, and David Heckerman
Microsoft Research
Redmond WA, 98052-6399
{thiesson,meek,dmax,heckerma}@microsoft.com

## Abstract

We describe computationally efficient methods for learning mixtures in which each component is a directed acyclic graphical model (mixtures of DAGs or MDAGs). We argue that simple search-and-score algorithms are infeasible for a variety of problems, and introduce a feasible approach in which parameter and structure search is interleaved and expected data is treated as real data. Our approach can be viewed as a combination of (1) the Cheeseman–Stutz asymptotic approximation for model posterior probability and (2) the Expectation–Maximization algorithm. We evaluate our procedure for selecting among MDAGs on synthetic and real examples.

## 1 Introduction

For almost a decade, statisticians and computer scientists have used directed-acyclic graph (DAG) models for learning from data (e.g., Cooper & Herskovits, 1992; Spirtes, Glymour, & Scheines, 1993; Spiegelhalter, Dawid, Lauritzen, & Cowell, 1993; Buntine, 1994; and Heckerman, Geiger, & Chickering, 1995). In this paper, we consider mixtures of DAG models (MDAG models) and methods for choosing among models in this class. MDAG models generalize DAG models, and should more accurately model domains containing multiple distinct populations. In general, our hope is that the use of MDAG models will lead to better predictions and more accurate insights into causal relationships. In this paper, we concentrate on prediction.

We take a decidedly Bayesian perspective on the problem of learning MDAG models. In principle, learning is straightforward: we compute the posterior probability of each model in the class given data and use this criterion to average over the models or to select one or more models. From a computational perspective, however, learning is extremely difficult. One problem is that the number of possible model structures grows super-exponentially with the number of random variables for the domain. A second problem is that all available methods for computing the posterior probability of an MDAG model, including Monte-Carlo and large-sample approximations, are slow. In combination, these problems make simple search-and-score learning algorithms intractable for MDAG models.

In the paper, we introduce a heuristic method for MDAG model selection that addresses both of these difficulties. The method is not guaranteed to find the MDAG model with the highest probability, but experiments that we present suggest that it often identifies a good one. Our approach handles missing data and component DAG models that contain hidden or latent variables. Our approach can be used to learn DAG models (single-component MDAG models) from incomplete data as well.

## 2 Multi-DAG models and mixtures of DAG models

In this section, we describe DAG, multi-DAG, and MDAG models. First, however, let us introduce some notation. We denote a random variable by an upper-case letter (e.g., $X, Y, X_i, \Theta$), and the value of a corresponding random variable by that same letter in lower case (e.g., $x, y, x_i, \theta$). When $X$ is discrete, we use $|X|$ to denote the number of values of $X$, and sometimes refer to a value of $X$ as a *state*. We denote a set of random variables by a bold-face capitalized letter or letters (e.g., $\mathbf{X}, \mathbf{Y}, \mathbf{Pa}_i$). We use a corresponding bold-face lower-case letter or letters (e.g., $\mathbf{x}, \mathbf{y}, \mathbf{pa}_i$) to denote an assignment of value to each random variable in a given set. When $\mathbf{X} = \mathbf{x}$ we say that $\mathbf{X}$ is in *configuration* $\mathbf{x}$. We use $p(\mathbf{X} = \mathbf{x}|\mathbf{Y} = \mathbf{y})$ (or $p(\mathbf{x}|\mathbf{y})$ as a shorthand) to denote the probability or probability density that $\mathbf{X} = \mathbf{x}$ given $\mathbf{Y} = \mathbf{y}$. We also use $p(\mathbf{x}|\mathbf{y})$ to denote the probability distribution (both mass functions and density functions) for $\mathbf{X}$ given $\mathbf{Y} = \mathbf{y}$. Whether $p(\mathbf{x}|\mathbf{y})$ refers to a probability, a probability density, or a probability distribution should be clear from context.

Suppose our problem domain consists of random variables $\mathbf{X} = (X_1, \ldots, X_n)$. A *DAG model* for $\mathbf{X}$ is a graphical factorization of the joint probability distribution of $\mathbf{X}$. The model consists of two components: a structure and a set of local distribution families. The structure $\mathbf{b}$ for $\mathbf{X}$ is a directed acyclic graph that represents conditional-independence assertions through a

factorization of the joint distribution for $\mathbf{X}$:

$$p(\mathbf{x}) = \prod_{i=1}^{n} p(x_i|\mathbf{pa}(\mathbf{b})_i) \qquad (1)$$

where $\mathbf{pa}(\mathbf{b})_i$ is the configuration of the parents of $X_i$ in structure $\mathbf{b}$ consistent with $\mathbf{x}$. The local distribution families associated with the DAG model are those in Equation 1. In this discussion, we assume that the local distribution families are parametric. Using $\boldsymbol{\theta}_b$ to denote the collective parameters for all local distributions, we rewrite Equation 1 as

$$p(\mathbf{x}|\boldsymbol{\theta}_b) = \prod_{i=1}^{n} p(x_i|\mathbf{pa}(\mathbf{b})_i, \boldsymbol{\theta}_b) \qquad (2)$$

With one exception to be discussed in Section 6, the parametric family corresponding to the variable $X$ will be determined by (1) whether $X$ is discrete or continuous and (2) the model structure. Consequently, we suppress the parametric family in our notation, and refer to the DAG model simply by its structure $\mathbf{b}$.

Let $\mathbf{b}^h$ denote the assertion or hypothesis that the "true" joint distribution can be represented by the DAG model $\mathbf{b}$ and has precisely the conditional independence assertions implied by $\mathbf{b}$. We find it useful to include the structure hypothesis explicitly in the factorization of the joint distribution when we compare model structures. In particular, we write

$$p(\mathbf{x}|\boldsymbol{\theta}_b, \mathbf{b}^h) = \prod_{i=1}^{n} p(x_i|\mathbf{pa}_i, \boldsymbol{\theta}_b, \mathbf{b}^h) \qquad (3)$$

This notation often makes it unnecessary to use the argument $\mathbf{b}$ in the term $\mathbf{pa}(\mathbf{b})_i$, and we use the simpler expression where possible.

The structure of a DAG model encodes a limited form of conditional independence that we call *context-non-specific conditional independence*. In particular, if the structure implies that two sets of random variables $\mathbf{Y}$ and $\mathbf{Z}$ are independent given some configuration of random variables $\mathbf{W}$, then $\mathbf{Y}$ and $\mathbf{Z}$ are also independent given every other configuration of $\mathbf{W}$. In a more general form of conditional independence, two sets of random variables may be independent given one configuration of $\mathbf{W}$, and dependent given another configuration of $\mathbf{W}$.

A multi-DAG model, called a Bayesian multinet by Geiger & Heckerman (1996), is a generalization of the DAG model that can encode context-specific conditional independence. In particular, *a multi-DAG model for $\mathbf{X}$ and distinguished random variable $C$* is a set of *component DAG models* for $\mathbf{X}$, each of which encodes the joint distribution for $\mathbf{X}$ given a state of $C$, and a distribution for $C$. Thus, the multi-DAG model for $\mathbf{X}$ and $C$ encodes a joint distribution for $\mathbf{X}$ and $C$, and can encode context-specific conditional independence among these random variables, because the structure of each component DAG model may be different.

Let $\mathbf{s}$ and $\boldsymbol{\theta}_s$ denote the structure and parameters of a multi-DAG model for $\mathbf{X}$ and $C$. In addition, let $\mathbf{b}_c$ and $\boldsymbol{\theta}_c$ denote the structure and parameters of the $c$th DAG-model component of the multi-DAG model. Also, let $\mathbf{s}^h$ denote the hypothesis that the "true" joint distribution for $\mathbf{X}$ and $C$ can be represented by the MDAG model $\mathbf{s}$ and has precisely the conditional independence assertions implied by $\mathbf{s}$. Then, the joint distribution for $\mathbf{X}$ and $C$ encoded by this multi-DAG model is given by

$$\begin{aligned} p(c, \mathbf{x}|\boldsymbol{\theta}_s, \mathbf{s}^h) &= p(c|\boldsymbol{\theta}_s, \mathbf{s}^h)\, p(\mathbf{x}|c, \boldsymbol{\theta}_s, \mathbf{s}^h) \\ &= \pi_c\, p(\mathbf{x}|\boldsymbol{\theta}_c, \mathbf{b}_c^h) \end{aligned} \qquad (4)$$

where $\boldsymbol{\theta}_s = (\boldsymbol{\theta}_1, \ldots, \boldsymbol{\theta}_{|C|}, \pi_1, \ldots, \pi_{|C|})$ are the parameters of the multi-DAG model, $\pi_c = p(c|\boldsymbol{\theta}_s, \mathbf{s}^h)$, and $\mathbf{b}_c^h$ is a shorthand for the conjunction of the events $\mathbf{s}^h$ and $C = c$. As with DAG models, we sometimes use the structure alone to refer to the multi-DAG model.

In what follows, we assume that the distinguished random variable has a multinomial distribution. In addition, with one exception to be discussed in Section 6, we limit the structure of the component DAG models and the parametric families for the local distributions as follows. When $X_i \in \mathbf{X}$ is a discrete random variable, we require that every random variable in $\mathbf{Pa}_i$ (for every component model) also be discrete, and that the local distribution families for $X$ be a set of multinomial distributions, one for each configuration of $\mathbf{Pa}_i$. When $X_i \in \mathbf{X}$ is a continuous random variable, we require that the local distribution family for $X_i$ be a set of linear-regressions over $X_i$'s continuous parents with Gaussian error, one regression for each configuration of $X_i$'s discrete parents. Lauritzen (1992) refers to this set of restrictions as a *conditional-Gaussian distribution for a DAG model*.

In this paper, we concentrate on the special case where the distinguished random variable $C$ is hidden. In this situation, we are interested in the joint distribution for $\mathbf{X}$, given by

$$p(\mathbf{x}|\boldsymbol{\theta}_s, \mathbf{s}^h) = \sum_{c=1}^{|C|} \pi_c\, p(\mathbf{x}|\boldsymbol{\theta}_c, \mathbf{b}_c^h) \qquad (5)$$

This joint distribution is a mixture of distributions determined by the component DAG models, and has mixture weights $\pi_1, \ldots, \pi_{|C|}$. Thus, when $C$ is hidden, we say that the multi-DAG model for $\mathbf{X}$ and $C$ is a *mixture of DAG models (or MDAG model) for $\mathbf{X}$*.

An important subclass of DAG models is the *Gaussian DAG model* (e.g., Shachter & Kenley, 1989). In this subclass, the local distribution family for every random variable given its parents is a linear regression with Gaussian noise. It is well known that a Gaussian DAG model for $X_1, \ldots, X_n$ uniquely determines a multivariate-Gaussian distribution for those random variables. In such a model, the structure of the DAG model (in part) determines the "shape" of the multivariate-Gaussian distribution. Thus, the MDAG model class includes mixtures of multivariate-Gaussian distributions in which each component may have a different shape.

## 3 Learning multi-DAG models

In this and the following two sections, we consider a Bayesian approach for learning multi-DAG models and MDAG models. Let us assume that our data is exchangeable so that we can reason as if the data is a random sample from a true joint distribution. In addition, let us assume that the true joint distribution for $\mathbf{X}$ is encoded by some multi-DAG model, and that we are uncertain about both its structure and parameters. We define a discrete random variable $\mathbf{S}^h$ whose states $\mathbf{s}^h$ correspond to the possible true model hypotheses, and encode our uncertainty about structure using the probability distribution $p(\mathbf{s}^h)$. In addition, for each model $\mathbf{s}$, we define a continuous vector-valued random variable $\Theta_s$, whose configurations $\boldsymbol{\theta}_s$ correspond to the possible true parameters. We encode our uncertainty about $\Theta_s$ using the probability density function $p(\boldsymbol{\theta}_s|\mathbf{s}^h)$.

Given a random sample $\mathbf{d} = (\mathbf{x}_1, \ldots, \mathbf{x}_N)$ from the true distribution for $\mathbf{X}$, we compute the posterior distributions for each $\mathbf{s}^h$ and $\boldsymbol{\theta}_s$ using Bayes' rule.

We can use the model posterior probability for various forms of model comparison, including model averaging (e.g., Bernardo & Smith, 1994). In this work, we limit ourselves to the selection of a model with a high posterior probability. In what follows, we concentrate on model selection using the posterior model probability. To simplify the discussion, we assume that all possible model structures are equally likely, a priori, in which case our selection criterion is the marginal likelihood:

$$p(\mathbf{d}|\mathbf{s}^h) = \int p(\mathbf{d}|\boldsymbol{\theta}_s, \mathbf{s}^h) \, p(\boldsymbol{\theta}_s|\mathbf{s}^h) \, d\boldsymbol{\theta}_s \quad (6)$$

### 3.1 The marginal likelihood criterion

Consider a DAG model $\mathbf{b}$ that encodes a conditional-Gaussian distribution for $\mathbf{X}$. Let $\Theta_i, i = 1, \ldots, n$ denote the random variables corresponding to the parameters of the local distribution family for $X_i$. Buntine (1994) and Heckerman and Geiger (1995) have shown that, if (1) the parameters $\Theta_1, \ldots, \Theta_n$ are mutually independent given $\mathbf{b}^h$, (2) the parameter priors $p(\Theta_i|\mathbf{b}^h)$ are conjugate for all $i$, and (3) the data $\mathbf{d}$ is complete for $C$ and $\mathbf{X}$, then the log marginal likelihood has a closed form that can be computed efficiently.

This observation extends to multi-DAG models. Let $\Theta_{ic}$ denote the set of random variables corresponding to the local distribution family of $X_i$ in component $c$. Also, let $\Pi$ denote the set of random variables $(\Pi_1, \ldots, \Pi_{|C|-1})$ corresponding to the mixture weights. If (1) $\Pi, \Theta_{11}, \ldots, \Theta_{n1}, \ldots, \Theta_{1|C|}, \ldots, \Theta_{n|C|}$ are mutually independent given $\mathbf{s}^h$, (2) the parameter priors $p(\Theta_{ic}|\mathbf{s}^h)$ are conjugate for all $i$ and $c$, and (3) the data $\mathbf{d}$ is complete, then the marginal likelihood $p(\mathbf{d}|\mathbf{s}^h)$ has a closed form. In particular,

$$\log p(\mathbf{d}|\mathbf{s}^h) = \log p(\mathbf{d}^C) + \sum_{c=1}^{|C|} \log p(\mathbf{d}^{\mathbf{X},C=c}|\mathbf{b}_c^h) \quad (7)$$

where $\mathbf{d}^C$ is the data restricted to the variable $C$, and $\mathbf{d}^{\mathbf{X},C=c}$ is the data restricted to the variables $\mathbf{X}$ and those cases in which $C = c$. The term $p(\mathbf{d}^C)$ is the marginal likelihood of a trivial DAG model having only a single discrete node $C$. The terms in the sum are log marginal likelihoods for the component DAG models of the multi-DAG. Hence, $p(\mathbf{d}|\mathbf{s}^h)$ has a closed form.

### 3.2 Structure search

An important issue regarding model selection is the search for models (structures) with high posterior probabilities. Consider the problem of finding the DAG model with the highest marginal likelihood from the set of all models in which each node has no more than $k$ parents. Chickering (1996) has shown the problem for $k > 1$ is NP-hard. It follows immediately that the problem of finding the multi-DAG model with the highest marginal likelihood from the set of all multi-DAGs in which each node in each component has no more than $k$ parents is NP-hard. Consequently, researchers use heuristic search algorithms including greedy search, greedy search with restarts, best-first search, and Monte-Carlo methods.

One consolation is that various model-selection criteria, including log marginal likelihood (under the assumptions just described), are factorable. We say that a criterion crit$(\mathbf{s}, \mathbf{d})$ for a multi-DAG structure $\mathbf{s}$ is *factorable* if it can be written as follows:

$$\text{crit}(\mathbf{s}, \mathbf{d}) = f(\mathbf{d}^C) + \sum_{c=1}^{|C|} \sum_{i=1}^{n} g_c(\mathbf{d}^{X_i, \mathbf{Pa}_i^c}) \quad (8)$$

where $\mathbf{d}^C$ is the data restricted to the set $C$, $\mathbf{Pa}_i^c$ are the parents of $X_i$ in component $c$, $\mathbf{d}^{X_i,\mathbf{Pa}_i^c}$ is the data restricted to the random variables $X_i$ and $\mathbf{Pa}_i^c$ and to those cases in which $C = c$, and $f$ and $g_c$ are functions. When a criterion is factorable, search is more efficient for two reasons. One, the component DAG models have non-interacting subcriteria so that we may search for a good DAG structure for each component separately. Two, as we search for a good structure in any one component, we need not reevaluate the criterion for the whole component. For example, in a greedy search for a good DAG structure, we iteratively transform the graph by choosing the transformation that improves the model criterion the most, until no such transformation is possible. Typical transformations include the removal, reversal, and addition of an arc (constrained so that the resulting graph is acyclic). Given a factorable criterion, we only need to reevaluate $g_c$ for $X_i$ if it's parents have changed.

## 4 Learning MDAGs: A simple approach

When learning multi-DAG models given complete data, the marginal likelihood has a closed form. In contrast, when learning MDAGs, the assumption that data is complete does not hold, because the distinguished random variable $C$ is hidden. When data is incomplete, no tractable closed form for marginal likelihood is available. Nonetheless, we can approximate the marginal likelihood using either Monte-Carlo or large-sample methods (e.g., DiCiccio, Kass, Raftery, and Wasserman, 1995). Thus, a straightforward class of algorithm for choosing an MDAG model is to search among structures as before (e.g., perform greedy search), using some approximation for marginal likelihood. We shall refer to this class as *simple search-and-score algorithms*.

As we shall see, simple search-and-score algorithms for MDAG model selection are computationally infeasible in practice. Nonetheless, let us consider one approximation for the marginal likelihood that will help motivate a tractable class of algorithms that we consider in the next section. The approximation that we examine is a large-sample approximation first proposed by Cheeseman & Stutz (1995):

$$p(\mathbf{d}|\mathbf{s}^h) \approx p(\mathbf{d}'|\mathbf{s}^h) \, \frac{p(\mathbf{d}'|\tilde{\boldsymbol{\theta}}_s, \mathbf{s}^h)}{p(\mathbf{d}|\tilde{\boldsymbol{\theta}}_s, \mathbf{s}^h)} \quad (9)$$

where $\mathbf{d}'$ is any completion of the data set $\mathbf{d}$.

The approximation is a heuristic one, but Chickering & Heckerman (1997) give an argument that it may perform well in practice[1]. Furthermore, they provide an empirical study, using multinomial mixtures, that shows the approximation to be quite good. In all experiments, it was at least as accurate and sometimes more accurate than the standard approximation obtained using Laplace's method (e.g., Tierney & Kadane, 1986).

An important idea behind the Cheeseman–Stutz approximation is that we treat data completed by the EM algorithm as if it were real data. This same idea underlies the M step of the EM algorithm. As we shall see in the next section, this idea also can be applied to structure search.

## 5 Learning MDAGs: A practical approach

Simple search-and-score algorithms for selecting MDAG models are inefficient for two reasons. One is that computing approximations for the marginal likelihood is slow (DiCiccio et al., 1995). Another is that these approximations do not factor. Consequently, every time a transformation is applied to a structure during search, the entire structure may need to be rescored. In this section, we consider a heuristic approach that addresses both of these problems.

---
[1]Chickering & Heckerman (1997) discuss a version of the Cheeseman–Stutz approximation that has a correction for dimension.

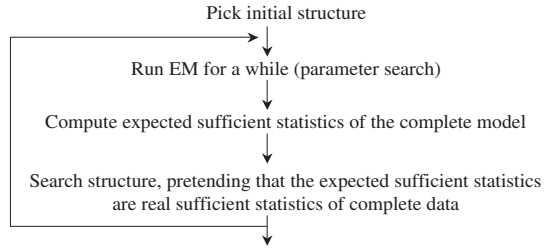

Figure 1: A schematic of our approach for MDAG model selection.

The basic idea behind the approach is that we interleave parameter search with structure search. A schematic of this approach is shown in Figure 1. First, we choose some initial model and parameter values. Then, we perform several iterations of the EM algorithm to find fairly good values for the parameters of the structure. Next, we use these parameter values and the current model to compute expected sufficient statistics for a *complete* MDAG (one that encodes no conditional-independence facts). We call these statistics for the current model $\mathbf{s}$, parameters $\boldsymbol{\theta}_s$, and data $\mathbf{d}$ the *expected complete model sufficient statistics* and denote the quantity by $\mathrm{ECMSS}(\mathbf{d}, \boldsymbol{\theta}_s, \mathbf{s})$. A detailed discussion of the computation of this quantity is given in the Appendix. Next, we treat these expected sufficient statistics as if they were sufficient statistics from a complete data set, and perform structure search. Because we pretend the data set is complete, the model scores have a closed form and are factorable, making structure search efficient. After structure search, we reestimate the parameters for the new structure to be the MAP parameters given the expected sufficient statistics. Finally, the EM, the $\mathrm{ECMSS}(\mathbf{d}, \boldsymbol{\theta}_s, \mathbf{s})$ computation, the structure search, and the parameter reestimation steps are iterated until some convergence criterion is satisfied.

In the remainder of this section, we discuss variations of the approach. In addition, we examine the criterion used for model search, the initialization of both the structure and parameters, and an approach for determining the number of mixture components and the number of states of any hidden variables in the component models.

Our search criterion is the log marginal likelihood of the expected complete model sufficient statistics:

$$\mathrm{crit}(\mathbf{s}'|\mathbf{d}, \tilde{\boldsymbol{\theta}}_s, \mathbf{s}) = \log p(\mathrm{ECMSS}(\mathbf{d}, \tilde{\boldsymbol{\theta}}_s, \mathbf{s})|\mathbf{s}'^h) \quad (10)$$

where $\mathbf{s}'$ is the model being evaluated and $(\mathbf{s}, \boldsymbol{\theta}_s)$ are the model and parameters used to compute the expected complete model sufficient statistics.

We use $(\mathbf{s}, \boldsymbol{\theta}_s)$ to compute sufficient statistics for the *complete* model, because we want all possible dependencies in the data to be reflected in the statistics. If we were to compute sufficient statistics for an incomplete (constrained) model, then models visited dur-ing search that violates these constraints would not be supported by the data.

The criterion in Equation 10 is related to the Cheeseman–Stutz approximation for the marginal likelihood, which we can rewrite as

$$\log p(\mathbf{d}|\mathbf{s}'^h) = \log p(\text{ECMSS}(\mathbf{d}, \tilde{\boldsymbol{\theta}}_{s'}, \mathbf{s}')|\mathbf{s}'^h)$$
$$+ \log \frac{p(\mathbf{d}|\tilde{\boldsymbol{\theta}}_{s'}, \mathbf{s}'^h)}{p(\text{ECMSS}(\mathbf{d}, \tilde{\boldsymbol{\theta}}_{s'}, \mathbf{s}')|\tilde{\boldsymbol{\theta}}_{s'}, \mathbf{s}'^h)}. (11)$$

Although the argument of Chickering & Heckerman (1997) suggests that Equation 11 is a more accurate approximation for the log marginal likelihood than is Equation 10, we use the less accurate criterion for two practical reasons. One, if we were to include the likelihood ratio "correction term" in Equation 11, then the criterion would not factor. Two, if we were to use just the first term in Equation 11, then we would still need to compute the MAP configuration $\tilde{\boldsymbol{\theta}}_{s'}$ for every structure that we evaluate. In contrast, by using Equation 10, we compute the MAP configuration $\tilde{\boldsymbol{\theta}}_s$ only once. Despite these shortcuts, experiments described in Section 6 suggest that the use of the criterion in Equation 10 guides the structure search to good models.

Our approach requires that both an initial structure and an initial parameterization be chosen. First, let us consider structural initialization. We initialize the structure of each component model by placing an arc from every hidden variable to every observable variable, with the exception that nodes corresponding to continuous random variables do not point to nodes corresponding to discrete random variables. A simpler choice for an initial graph is one in which every component consists of an empty graph—that is, a graph containing no arcs. However, with such an initialization and for a restricted set of priors, we conjecture that our approach would be unable to discover connections between hidden and observable variables.

Next, let us consider parameter initialization. When the mixture components contain no hidden continuous variables, we initialize parameters for a component DAG structure $\mathbf{b}$ as follows. First, we remove all hidden nodes and adjacent arcs from $\mathbf{b}$, creating model $\mathbf{b}'$. Next, we determine $\tilde{\boldsymbol{\theta}}_{b'}$, the MAP configuration for $\boldsymbol{\theta}_{b'}$ given data $\mathbf{d}$. Since the data is complete with respect to $\mathbf{b}'$, we can compute this MAP in closed form. Then, we create a conjugate distribution for $\boldsymbol{\theta}_{b'}$ whose configuration of maximum value agrees with the MAP configuration just computed and whose equivalent sample sizes are specified by the user. Next, for each non-hidden node $X_i$ in $\mathbf{b}$ and for each configuration of $X_i$'s hidden discrete parents, we initialize the parameters of the local distribution family for $X_i$ by drawing from the conjugate distribution just described. For each hidden discrete node $X_i$ in $\mathbf{b}$ and for each configuration of $X_i$'s (possible) parents, we initialize the multinomial parameters associated with the local distribution family of $X_i$ to be some fixed distribution (e.g., uniform). When the mixture components contain hidden continuous variables, we use the simpler approach of initializing parameters at random (i.e., by drawing from a distribution such as the prior). Methods for initializing the parameters of the distinguished random variable $C$ include (1) setting the parameters to be equal, (2) setting the parameters to their prior means, and (3) drawing the parameters from a Dirichlet distribution.

As we have mentioned, our approach has several variations. One source of variation is the heuristic algorithm used for search once ECMSS$(\mathbf{d}, \boldsymbol{\theta}_s, \mathbf{s}))$ is computed. The options are the same as those for the simple search-and-score algorithms, and include greedy search, greedy search with restarts, best-first search, and Monte-Carlo methods. In preliminary studies, we have found greedy search to be effective; and in our analysis of real data in Section 6, we use this technique.

Another source of variation is the schedule used to alternate between parameter and structure search. With respect to parameter search, we can run EM to convergence, for one step, for some fixed number of steps, or for a number of steps that depends on how many times we have performed the search phase. With respect to structure search, we can perform model-structure transformations for some fixed number of steps, for some number of steps that depends on how many times we have performed the search phase, or until we find a local maximum. Finally, we can iterate the steps consisting of EM, the computation of ECMSS$(\mathbf{d}, \boldsymbol{\theta}_s, \mathbf{s})$, and structure search until either (1) the MDAG structure does not change across two consecutive search phases, or (2) the approximate marginal likelihood of the resulting MDAG structure does not increase. Under the second schedule, the algorithm is guaranteed to terminate, because the marginal likelihood cannot increase indefinitely. Under the first schedule, we do not know of a proof that the algorithm will terminate. In our experiments with greedy structure search, however, we have found that this schedule halts.

We find it convenient to describe these schedules using a regular grammar, where E, M, $E_c$, S denote an E step, M step, computation of ECMSS$(\mathbf{d}, \boldsymbol{\theta}_s, \mathbf{s})$, and structure search, respectively. For example, we use $((EM)^*E_cS^*M)^*$ to denote the case where, within each outer iteration, we (1) run EM to convergence, (2) compute the expected complete model sufficient statistics, (3) run structure search to convergence, and (4) perform an M step. Another schedule we examine is $((EM)^{10}E_cS^*M)^*$. In this schedule, we run EM for only 10 steps before computing the expected complete model sufficient statistics.[2]

In a technical report that is a companion to this paper (Thiesson, Meek, Chickering, and Heckerman, 1997), we evaluate various combinations of these schedules.

---

[2]When the structure search leaves the model structure unchanged, we force another iteration of the outer loop in which we run EM to convergence rather than for 10 steps. If the model structure changes in this forced iteration, we continue to iterate with 10 EM steps.

Our experiments indicate that, although it is not necessary to run EM to convergence between structure search, a single EM step between structure searches selects models that have lower prediction accuracy. We have found that the schedule $((EM)^{10}E_cS^*M)^*$ works well for a variety of problems.

Finally, the algorithm as described can compare neither models that contain different random variables nor models in which the same random variable has a different number of states. Nonetheless, we can perform an additional search over the number of states of each discrete hidden variable by applying the algorithm in Figure 1 to initial models with different numbers of states for the hidden variables. We can discard a discrete hidden variable from a model by setting its number of states to one. After the best MDAG for each initialization is identified, we select the overall best structure using some criterion. Because only a relatively small number of alternatives are considered, we can use a computationally expensive approximation for the marginal likelihood such as the Cheeseman-Stutz approximation or a Monte-Carlo method.

## 6 Example

In this section, we evaluate the predictive performance of MDAG models on real data. In addition, we evaluate some of the assumptions underlying our method for learning these models. In the domain that we consider, all the observable random variables are continuous. Consequently, we focus our attention on mixtures of Gaussian DAG models. To accommodate the outliers contained in the data set that we analyze, each of the mixture models that we consider has a noise component in addition to one or more Gaussian components. The noise component is modeled as a multivariate uniform distribution, and can be viewed as an empty DAG model in which the distribution function for each of the random variables is uniform.

We compare the predictive performance of (1) mixtures of DAG models (MDAG/n) (2) mixtures of multivariate-Gaussian distributions for which the covariance matrices are diagonal (MDIAG/n), and (3) mixtures of multivariate-Gaussian distributions for which the covariance matrices are full (MFULL/n). The MDIAG/n and MFULL/n model classes correspond to MDAG models with fixed empty structures and fixed complete structures, respectively, for all Gaussian components. The /n suffix indicates the existence of a uniform noise component.

We perform an outer search to identify the number of components within each mixture model as described in Section 5. In particular, we first learn a two-component model (one Gaussian and one noise component), and then increase by one the number of Gaussian mixture components until the model score is clearly decreasing. We choose the best number of components using the Cheeseman–Stutz criterion. Then, we measure the predictive ability of the chosen model s using the logarithmic scoring rule of Good (1952):

$$\frac{1}{|\mathbf{d}_{\text{test}}|} \sum_{l \in \mathbf{d}_{\text{test}}} \log p(\mathbf{x}_l | \mathbf{s}^h) \qquad (12)$$

where $\mathbf{d}_{\text{test}}$ is a set of test cases and $|\mathbf{d}_{\text{test}}|$ is the number of test cases. We approximate $p(\mathbf{x}_l|\mathbf{s}^h)$ by $p(\mathbf{x}_l|\tilde{\boldsymbol{\theta}}_s, \mathbf{s}^h)$, the likelihood evaluated at the MAP parameter configuration.[3]

When learning MDAG/n models, we use the $((EM)^{10}E_cS^*M)^*$ search schedule; and when learning MDIAG/n and MFULL/n models, we run the EM algorithm to convergence. In all experiments, we deem EM to have converged when the the ratio of the change in log likelihood from the proceeding step and the change in log likelihood from the initialization falls below $10^{-6}$. We initialize structure and parameters for our search procedures as described in Section 5 with equivalent sample sizes equal to 200.

The example we consider addresses the digital encoding of handwritten digits (Hinton, Dayan, & Revow, 1997). In this domain, there are 64 random variables corresponding to the gray-scale values [0,255] of scaled and smoothed 8-pixel x 8-pixel images of handwritten digits obtained from the CEDAR U.S. postal service database (Hull, 1994). Applications of joint prediction include image compression and digit classification. The sample sizes for the digits ("0" through "9") range from 1293 to 1534. For each digit, we use 1100 samples for training, and the remaining samples for testing.

We use a relatively diffuse Normal-Wishart parameter prior for each of the Gaussian components of MDIAG/n and MFULL/n models. In the notation of DeGroot (1970), our prior has $\nu = 2$, all values in $\boldsymbol{\mu}$ set to 64 as a rough assessment of the average gray-scale value over pixels, $\alpha = \nu + 64$, and $\boldsymbol{\tau}$ set to the identity matrix. We choose $\alpha$ to be the sum of $\nu$ and the number of observed variables to compute the MAP parameter values in closed form. The parameter priors for the Gaussian components of the MDAG/n models are Normal-Wishart priors specified using the hyperparameters described above and the methods described in Heckerman and Geiger (1995). We use a uniform prior on the number of components in the mixture and, when learning MDAG/n models, a uniform prior on the structure of the component DAG models. Because we know that the values of each of the 64 variables are constrained to the range [0,255], we fix the parameters in the uniform distribution of the noise model accordingly. Finally, the hyperparameters $\{\alpha_0, \ldots, \alpha_k\}$ of the Dirichlet prior on the mixture weights (i.e., the distinguished variable) are $\alpha_0 = 0.01$ for the noise component, and $\alpha_1 = \ldots = \alpha_k = 0.99/k$ for the Gaussian components.

The predictive logarithmic score on the test set for each digit is shown in Figure 2. The number of Gaussian components $k$ and the model dimension $d$ for the best model in each class are displayed in Table 1. Figure 2 indicates that MDAG/n models, on average, improve the predictive accuracy by 8% over MFULL/n models and 20% over MDIAG/n models. Note that the gains in predictive accuracy over MFULL/n models are obtained while reducing the average number of parameters by one third.

---

[3]We are currently implementing a Monte-Carlo method to average over the parameter configurations.

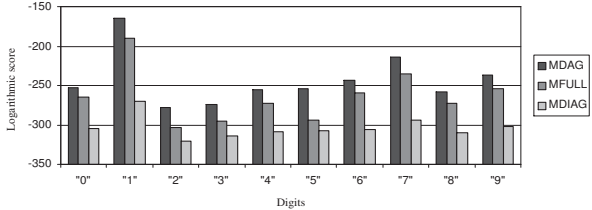

Figure 2: Logarithmic predictive scores on the test sets for the digit data.

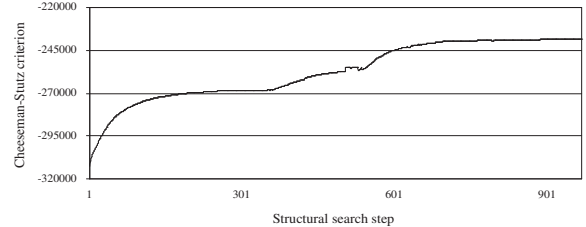

Figure 4: The Cheeseman–Stutz criterion for each intermediate model obtained during structure search when learning a three-component model for the digit "7". The abrupt increases around steps 1, 350, and 540 occur when structure search transitions to a new component.

| Digit | MDAG/n $k$ | MDAG/n $d$ | MFULL/n $k$ | MFULL/n $d$ | MDIAG/n $k$ | MDIAG/n $d$ |
|---|---|---|---|---|---|---|
| "0" | 5 | 1812 | 2 | 4290 | 8 | 1032 |
| "1" | 7 | 2910 | 2 | 4290 | 5 | 645 |
| "2" | 6 | 1816 | 1 | 2145 | 6 | 774 |
| "3" | 4 | 1344 | 1 | 2145 | 6 | 774 |
| "4" | 7 | 2115 | 2 | 4290 | 6 | 774 |
| "5" | 9 | 2702 | 1 | 2145 | 6 | 774 |
| "6" | 9 | 2712 | 2 | 4290 | 5 | 645 |
| "7" | 9 | 3168 | 2 | 4290 | 4 | 516 |
| "8" | 6 | 1868 | 2 | 4290 | 5 | 645 |
| "9" | 8 | 2955 | 2 | 4290 | 5 | 645 |

Table 1: Number of Gaussian components and parameters in the learned models for the digit data.

Using a P6 200MHz computer, the time taken to learn the MDAG/n, MFULL/n, and MDIAG/n models for a single digit—including the time needed to find the optimal number of components—is, on average, 6.0, 1.5, and 1.9 hours, respectively. These times could be improved by using a more clever search for the optimal number of mixture components.

To better understand the differences in the distributions that these mixture models represent, we examine the individual Gaussian components for the learned MDAG/n, MFULL/n, and MDIAG/n models for the digit "7". The first row of Figure 3 shows the means for each of the components of each of the models. The mean values for the variables in each component are displayed in an 8 x 8 grid in which the shade of grey indicates the value of the mean. The displays indicate that each of the components of each type of model are capturing distinctive types of sevens. They do not, however, reveal any of the dependency structure in the component models. To help visualize these dependencies, we drew four samples from each component for each type of model. The grid for each sample is shaded to indicate the sampled values. Whereas the samples from the MDIAG/n components do look like sevens, they are mottled. This is not surprising, because each of the variables are conditionally independent given the component. The samples for the MFULL/n components are not mottled, but indicate that multiple types of sevens are being represented in one component. That is, several of the samples look blurred and appear to have multiple sevens superimposed. Generally, samples from each MDAG/n component look like sevens of the same distinct style, all of which closely resemble the mean.

Let us turn our attention to the evaluation of one of the key assumptions underlying our learning method. As we have discussed, the criterion used to guide structure search (Equation 10) is only a heuristic approximation to the true model posterior. To investigate the quality of this approximation, we can evaluate the model posterior using the Cheeseman-Stutz approximation (what we believe to be a more accurate approximation) for intermediate models visited during structure search. If the heuristic criterion is good, then the Cheeseman–Stutz criterion should increase as structure search progresses. We perform this evaluation when learning a three-component MDAG model for the digit "7" using the $((EM)^{10}E_cS^*M)^*$ schedule. For 149 out of the 964 model transitions, the Cheeseman–Stutz approximation decreased. Overall, however, as shown in Figure 4, the Cheeseman–Stutz score progresses upward to apparent convergence. We obtain similar results for other data sets. These results suggest that the heuristic criterion (Equation 10) is a useful guide for structure search.

Using statistics from this same experiment, we are able to estimate the time it would take to learn the MDAG model using the simple search-and-score approach described in Section 4. We find that the time to learn the three-component MDAG model for the digit "7", using the Cheeseman–Stutz approximation for model comparison, is approximately 6 years on a P6 200MHz computer, thus substantiating our previous claim about the intractability of simple search-and-score approaches.

Finally, a natural question is whether the Cheeseman–Stutz approximation for the marginal likelihood is accurate for model selection. The answer is important, because the MDAG models we select and evaluate are chosen using this approximation. Some evidence for the reasonableness of the approximation is provided by the fact that, as we vary the number of components of the MDAG models, the Cheeseman-Stutz and predic-

|       | MDAG/n | | | | | | | | | MFULL/n | | MDIAG/n | | | |
|-------|--------|--|--|--|--|--|--|--|--|---------|--|--------|--|--|--|
| Weight | 0.15 | 0.13 | 0.03 | 0.04 | 0.12 | 0.16 | 0.33 | 0.002 | 0.04 | 0.65 | 0.35 | 0.31 | 0.23 | 0.19 | 0.23 |
| Mean | | | | | | | | | | | | | | | |
| Samples | | | | | | | | | | | | | | | |

Figure 3: Means and samples from the components of the learned MDAG/n, MFULL/n, and MDIAG/n models for digit "7".

## 7 Structure learning: A preliminary study

As we have mentioned in the introduction, many computer scientists and statisticians are using statistical-inference techniques to learn the structure of DAG models from observational (i.e., non-experimental data). Pearl & Verma (1991) and Spirtes et al. (1993) have argued that, under a set of simple (and sometimes reasonable) assumptions, the structures so learned can be used to infer cause-and-effect relationships. An interesting possibility is that these results can be generalized so that we may use the structure of learned MDAG models to infer causal relationships in mixed populations (populations in which subgroups have different causal relationships). In this section, we present a preliminary investigation into how well our approach can learn MDAG structure.

We perform our analysis as follows. First, we construct a "gold-standard" MDAG model, and use the model to generate data sets of varying size. Then, for each data set, we use our approach to learn an MDAG model (without a noise component). Finally, we compare the structure of the learned model to that of the gold-standard model, and measure the minimum number of arc manipulations (additions, deletions, and reversals) needed to transform each learned component structure to the corresponding gold-standard structure.

The gold-standard model is an MDAG model for five continuous random variables. The model has three mixture components. The structure of the first and third components (COMP1 and COMP3) are identical and this structure is shown in Figure 5a. The structure of the second component (COMP2) is shown in Figure 5b. The DAGs are parameterized so that there is some spatial overlap. In particular, all unconditional means in COMP1 and COMP2 are zero; all means in COMP3 are equal to five; and all linear coefficients and conditional variances are one (see Shachter & Kenley, 1989).

We construct a data set of size $N = 3000$ by sampling

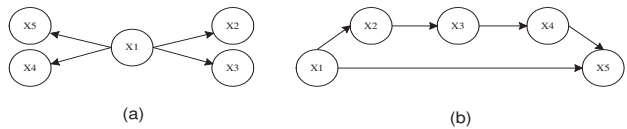

Figure 5: (a) The graphical structure for first and third components in the gold-standard MDAG. (b) The graphical structure for second component.

| Sample | | Weight of three | Arc differences | | |
|--------|---|-----------------|-------|-------|-------|
| size | $k$ | largest comp. | COMP1 | COMP2 | COMP3 |
| 93 | 2 | 1.00 | - | 4 | 0 |
| 186 | 2 | 1.00 | - | 2 | 0 |
| 375 | 3 | 1.00 | 1 | 1 | 0 |
| 750 | 5 | 0.98 | 1 | 1 | 0 |
| 1500 | 3 | 1.00 | 0 | 3 | 0 |
| 3000 | 5 | 0.99 | 1 | 1 | 0 |

Table 2: Performance on the task of structure learning as a function of sample size.

1000 cases from each component of the gold-standard model. We then iteratively subsample this data, creating data sets of size $N = 93, 186, 375, 750, 1500,$ and 3000.

Table 2 shows the results of learning models from the six data sets using the $((EM)^{10}E_cS^*M)^*$ schedule. The columns of the table contain the number of components $k$ in the learned MDAG, the sum of the mixture weights in the three largest components and the minimum number of arc manipulations (additions, deletions, and reversals) needed to transform each learned component structure to the corresponding gold-standard structure for the three components with the largest mixture weights. Arc manipulations that lead to a model with different structures but the same family of distributions are not included in the count. All learned MDAG structures are close to that of the gold-standard model. In addition, although not apparent from the table, the structure of every learned component has only additional arcs in comparison with the gold-standard model for sample sizes larger than 375. Finally, it is interesting to note that, essentially, the structure is recovered for a sample size as low as 375.

# 8 Related work

DAG models (single-component MDAG models) with hidden variables generalize many well-known statistical models including linear factor analysis, latent factor models (e.g., Clogg, 1995), and probabilistic principle component analysis (Tipping & Bishop, 1997). MDAG models generalize a variety of mixtures models including naive-Bayes models used for clustering (e.g., Clogg, 1995; Cheeseman and Stutz, 1995), mixtures of factor analytic models (Hinton, Dayan, & Revow, 1997), and mixtures of probabilistic principle component analytic models (Tipping & Bishop, 1997).

There is also work related to our learning methods. The idea of interleaving parameter and structure search to learn graphical models has been discussed by Meilă, Jordan, & Morris (1997), Singh (1997), and Friedman (1997). Meilă et al. (1997) consider the problem of learning mixtures of DAG models for discrete random variables where each component is a spanning tree. Similar to our approach, they treat expected data as real data to produce a completed data set for structure search. Unlike our work, they replace heuristic model search with a polynomial algorithm for finding the "best" spanning-tree components given the completed data. Also, unlike our work, they use likelihood as a selection criterion, and thus do not take into account the complexity of the model.

Singh (1997) concentrates on learning a single DAG model for discrete random variables with incomplete data. He does not consider continuous variables or mixtures of DAG models. In contrast to our approach, Singh (1997) uses a Monte-Carlo method to produce completed data sets for structure search.

Friedman (1997, 1998) describes general algorithms for learning DAG models given incomplete data, and provides theoretical justification for some of his methods. Similar to our approach and the approach of Meilă et al. (1997), Friedman treats expected data as real data to produce completed data sets. In contrast to our approach, Friedman obtains the expected sufficient statistics for a new model using the current model. Most of these statistics are calculated by performing probabilistic inference in the current DAG model, although some of the statistics are obtained from a cache of previous inferences. In our approach, we only need to perform inference once on every case that has missing values to compute the expected complete model sufficient statistics. After these statistics are computed, model scores for arbitrary structures can be computed without additional inference.

# 9 Discussion and future work

We have described mixtures of DAG models, a class of models that is more general than DAG models, and have presented a novel heuristic method for choosing good models in this class. Although evaluations for more examples (especially ones containing discrete variables) are needed, our preliminary evaluations suggest that model selection within this expanded model class can lead to substantially improved predictions. This result is fortunate, as our evaluations also show that simple search-and-score algorithms, in which models are evaluated one at a time using Monte-Carlo or large-sample approximations for model posterior probability, are intractable for some real problems.

One important observation from our evaluations is that the (practical) selection criterion that we introduce—the marginal likelihood of the complete-model sufficient statistics—is a good guide for model search. An interesting question is: Why? We hope that this work will stimulate theoretical work to answer this question and perhaps uncover better approximations for guiding model search. Friedman (1998) has some initial insight.

A possibly related challenge for theoretical study has to do with the apparent accuracy of the Cheeseman–Stutz approximation for the marginal likelihood. As we have discussed, in experiments with multinomial mixtures, Chickering & Heckerman (1997) have found the approximation to be at least as accurate and sometimes more accurate than the standard Laplace approximation. Our evaluations have also provided some evidence that the Cheeseman–Stutz approximation is an accurate criterion for model selection.

In our evaluations, we have not considered situations where the component DAG models themselves contain hidden variables. In order to learn models in this class, methods for structure search are needed. In such situations, the number of possible models is significantly larger than the number of possible DAGs over a fixed set of variables. Without constraining the set of possible models with hidden variables—for instance, by restricting the number of hidden variables—the number of possible models is infinite. On a positive note, Spirtes et al. (1993) have shown that constraint-based methods under suitable assumptions can sometimes indicate the existence of a hidden common cause between two variables. Thus, it may be possible to use the constraint-based methods to suggest an initial set of plausible models containing hidden variables that can then be subjected to a Bayesian analysis.

In Section 7, we saw that we can recover the structure of an MDAG model to a fair degree of accuracy. This observation raises the intriguing possibility that we can infer causal relationships from a population consisting of subgroups governed by different causal relationships. One important issue that needs to be addressed first, however, has to do with structural identifiability. For example, two MDAG models may superficially have different structures, but may otherwise be statistically equivalent. Although the criteria for structural identifiability among single-component DAG models is well known, such criteria are not well understood for MDAG models.

# Appendix: Expected complete model sufficient statistics

In this appendix, we examine complete model sufficient statistics more closely. We shall limit our discussion to multi-DAG models for which the component DAG models have conditional-Gaussian distributions. The extension to the noise component is straightforward.

Consider a multi-DAG model for random variables $C$ and $\mathbf{X}$. Let $\mathbf{\Gamma}$ denote the set of continuous variables in $\mathbf{X}$, $\boldsymbol{\gamma}$ denote a configuration of $\mathbf{\Gamma}$, and $n_c$ denote the number of variables in $\mathbf{\Gamma}$. Let $\mathbf{\Delta}$ denote the set of all discrete variables (including the distinguished variable $C$), and $m$ denote the number of possible configurations of $\mathbf{\Delta}$. In addition, let $\mathbf{d} = \mathbf{y}_1, \ldots, \mathbf{y}_N$, where $\mathbf{y}_i$ is the configuration of the observed variables in case $i$. Note that different variables may be observed in different cases. Finally, as in Dempster et al. (1977), let $\mathbf{x}_i$ denote the $i$th *complete case*—the configuration of $\mathbf{X}$ and $C$ in the $i$th case.

Now, consider the complete model sufficient statistics for a complete case, which we denote $T(\mathbf{x})$. For the multi-DAG model, $T(\mathbf{x})$ is a vector $\langle\langle N_1, R_1, S_1\rangle, \ldots, \langle N_m, R_m, S_m\rangle\rangle$ of $m$ triples, where the $N_j$ are scalars, the $R_j$ are vectors of length $n_c$, the $S_j$ are square matrices of size $n_c \times n_c$. In particular, if the discrete variables in $\mathbf{x}$ take on the $j^{th}$ configuration, then $N_j = 1$, $R_j = \boldsymbol{\gamma}$, and $S_j = \boldsymbol{\gamma}' * \boldsymbol{\gamma}$, and $N_k = 0$, $R_k = 0$, and $S_k = 0$ for $k \neq j$.

When we do not have a complete data set, we compute the expected complete model sufficient statistics $\text{ECMSS}(\mathbf{d}, \boldsymbol{\theta}_s, \mathbf{s})$, given by

$$\text{ECMSS}(\mathbf{d}, \boldsymbol{\theta}_s, \mathbf{s}) = \sum_{i=1}^{N} E(T(\mathbf{x}_i)|\mathbf{y}_i, \boldsymbol{\theta}_s, \mathbf{s}^h) \qquad (13)$$

The expectation is taken with respect to the joint distribution over the random variables $C$ and $\mathbf{X}$ given $\boldsymbol{\theta}_s$, $\mathbf{s}^h$, and the observations for the current case. The expectation of $T(\mathbf{x})$ is computed by performing probabilistic inference in the multi-DAG model. Such inference is a simple extension of the work of Lauritzen (1992). The sum of expectations are simply scalar, vector, or matrix additions (as appropriate) in each triple in each of the coordinates of the vector.

Note that, in the computation as we have described it, we require a statistic triple for every possible configuration of discrete variables. In practice, however, we can use a sparse representation in which we store triples only for those complete observations that are consistent with the data.